\documentclass{article}
\usepackage{microtype}
\usepackage{graphicx}
\usepackage{booktabs}
\usepackage{multirow}
\usepackage{amsmath,amssymb,amsfonts,mathtools}
\usepackage{xcolor}
\usepackage{wasysym}
\usepackage{hyperref}

\usepackage[acronym]{glossaries}

\usepackage[accepted]{mlsys2025_arxiv} 
\makeatletter
\renewcommand{\Notice@String}{Preprint. Under review at the Conference on Machine Learning and Systems (MLSys 2027)}
\makeatother

\usepackage{enumitem}

\setlist{
    nosep,
    topsep=2pt,
    partopsep=0pt,
    itemsep=1pt,
    parsep=0pt
}

\newcommand{\ind}{\mathbf{1}}

\mlsystitlerunning{Smart Content Ingestion for Generative AI Workloads}

\makeglossaries

\definecolor{tagE}{HTML}{1F4E79}
\definecolor{tagD}{HTML}{2A9D8F}
\definecolor{tagP}{HTML}{2E75B6}
\definecolor{tagH}{HTML}{E07A2F}
\newcommand{\E}{\textcolor{tagE}{\textbf{[E]}}}
\newcommand{\D}{\textcolor{tagD}{\textbf{[D]}}}
\newcommand{\Pp}{\textcolor{tagP}{\textbf{[P]}}}
\newcommand{\Hh}{\textcolor{tagH}{\textbf{[H]}}}
\newcommand{\code}[1]{\texttt{\small #1}}

\begin{document}

\newacronym{Al-Jazeera}{Al-Jazeera}{Al-Jazeera Satellite Channel}
\newacronym{AraGPT-2}{AraGPT-2}{Arabic Generative Pre-trained Transformer 2}
\newacronym{AraGPT-3}{AraGPT-3}{Arabic Generative Pre-trained Transformer 3}
\newacronym{ArSAS}{ArSAS}{Arabic Speech-Act and Sentiment}
\newacronym{PATB}{PATB}{Penn Arabic Treebank}
\newacronym{BBC}{BBC}{British Broadcasting Corporation}
\newacronym{BERT}{BERT}{Bidirectional Encoder Representations from Transformers}
\newacronym{BPE}{BPE}{byte-pair encoding}
\newacronym{mBERT}{mBERT}{multi-lingual BERT}
\newacronym{RoBERTa}{RoBERTa}{A Robustly Optimized BERT Pre-training Approach}
\newacronym{CA}{CA}{Classical Arabic}
\newacronym{CC}{CC}{Common Crawl}
\newacronym{CC-100}{CC-100}{Common Crawl 100}
\newacronym{CNN}{CNN}{Cable News Network}
\newacronym{DA}{DA}{Dialectal Arabic}
\newacronym{EAPCOUNT}{EAPCOUNT}{English-Arabic Parallel Corpus of the United Nations Texts}
\newacronym{GenAI}{GenAI}{Generative AI}
\newacronym{eVal}{eVal}{EnterpriseVal}
\newacronym{ML}{ML}{Machine Learning}
\newacronym{GPT-1}{GPT-1}{Generative Pre-trained Transformer 1}
\newacronym{GPT-2}{GPT-2}{Generative Pre-trained Transformer 2}
\newacronym{GPT-3}{GPT-3}{Generative Pre-trained Transformer 3}
\newacronym{GPT}{GPT}{Generative Pre-trained Transformer}
\newacronym{GPU}{GPU}{Graphics Processing Unit}
\newacronym{HPL}{HPL}{High-Performance Linpack} 
\newacronym{IR}{IR}{Information Retrieval} 
\newacronym{KSUCCA}{KSUCCA}{King Saud University Corpus of Classical Arabic}
\newacronym{LM}{LM}{Language model}
\newacronym{LLM}{LLM}{Large Language Model}
\newacronym{LSTM}{LSTM}{Long-Short Term Memory}
\newacronym{ME}{ME}{Middle East}
\newacronym{MGB-2}{MGB-2}{Multi-Genre Broadcast-2}
\newacronym{MSA}{MSA}{Modern Standard Arabic}
\newacronym{MEGLN}{MEGLN}{MSA, Egyptian, Gulf, Levant \& North African}
\newacronym{MRC}{MRC}{Machine Reading Comprehension}
\newacronym{MT}{MT}{Machine Translation}
\newacronym{NN}{NN}{Neural Network}
\newacronym{NQ}{NQ}{Natural Questions}
\newacronym{NLP}{NLP}{Natural Language Processing}
\newacronym{OpenITI}{OpenITI}{Open Islamicate Texts Initiative}
\newacronym{OSAC}{OSAC}{Open-Source Arabic Corpora}
\newacronym{OSCAR}{OSCAR}{Open Super-large Crawled ALMAnaCH coRpus}
\newacronym{OSIAN}{OSIAN}{Open Source International Arabic News}
\newacronym{POS}{POS}{Part-of-speech}
\newacronym{PTB}{PTB}{Penn Tree Bank}
\newacronym{QA}{QA}{Question Answering} 
\newacronym{RNN}{RNN}{Recurrent Neural Network}
\newacronym{RT}{RT}{Russia Today}
\newacronym{SM}{SM}{Sentence Match}
\newacronym{SQuAD}{SQuAD}{Stanford Question Answering Dataset}
\newacronym{TPU}{TPU}{Tensor Processing Unit}
\newacronym{TFRC}{TFRC}{TensorFlow Research Cloud}
\newacronym{KSU}{KSU}{King Saud University}
\newacronym{WER}{WER}{Word Error Rate}
\newacronym{XNLI}{XNLI}{Cross-lingual Natural Language Inference}
\newacronym{HARD}{HARD}{Hotel Arabic Reviews Dataset}

\twocolumn[
\mlsystitle{Smart Content Ingestion for Generative AI Workloads}

\mlsyssetsymbol{equal}{*}
\begin{mlsysauthorlist}
\mlsysauthor{Abbas Raza Ali}{equal,abbas}
\mlsysauthor{Muhammad Ajmal Siddiqui}{ajmal}
\mlsysauthor{Moona Zahid}{moona}
\end{mlsysauthorlist}

\mlsysaffiliation{abbas}{Citigroup Inc., London, United Kingdom}
\mlsysaffiliation{ajmal}{Ernst \& Young LLP, London, United Kingdom}
\mlsysaffiliation{moona}{NVIDIA Corporation, London, United Kingdom}

\mlsyscorrespondingauthor{Abbas Raza Ali}{abbas.raza.ali@gmail.com}
\mlsyskeywords{Generative AI, Document Ingestion, OCR, Chunking, Evaluation}

\vskip 0.3in

\begin{abstract}
The evolution of machine learning has progressively changed where intelligence resides in an AI system. In conventional \gls{ML} the task, data representation, labels and model architecture were tightly coupled, so data preparation was narrow, schema-bound and visible. Generative AI decouples the model from any single task: one foundation model serves open-ended downstream tasks, and the generality gained on the model side is matched by heterogeneity on the data side, because enterprise knowledge is authored in the formats people use (PDFs, presentations, spreadsheets, scanned documents, forms, tables, diagrams and mixed-layout files) that carry textual, visual, geometric and structural information at once. A language model or retriever cannot reason reliably over information that has not been represented correctly at this interface, so content extraction becomes a lifecycle stage in its own right whose errors no downstream retriever or re-ranker can repair. This paper presents a production-ready content-extraction system that makes this stage explicit, configurable, and measurable. The system incorporates selective OCR routing, a scarcity-first curation engine with a reference-based extraction scorer that measures character, word, and table-structure accuracy, a deterministic structure-aware parent-child chunker, and a read-only retrieval evaluator that generates grounded questions from every page and reports hit@$k$, mean reciprocal rank, and latency. On a 180-document corpus the best extractor scores 97.4 of 100 (character error rate 0.13\%, table similarity 0.995) and the chunker reaches hit@1 of 68.6\%, hit@10 of 92.8\% and MRR 0.77 over 25,050 generated questions. We distil three design principles (structure before semantics, never mutate what you measure, budget your labels) and position measured content extraction as the perception layer of enterprise agentic systems.
\end{abstract}
]

\printAffiliationsAndNotice{}

\section{Introduction}
\label{sec:intro}
The evolution of machine learning has progressively changed where intelligence resides in an AI system. In conventional \gls{ML} applications, the task, data representation, labels, and model architecture were closely coupled: a system was constructed for a particular prediction problem, and its training and evaluation data were designed around that objective. The emergence of \gls{GenAI} introduced a different systems paradigm where the models are trained on broad data at scale and are designed to support a wide range of downstream tasks, while \glspl{LLM} demonstrated that new tasks can often be specified through natural-language instructions and examples without updating model parameters \citep{Abbas2024}. The consequence is not simply a larger model: task specification increasingly moves from model construction into inference-time context, instructions, tools, and retrieved information.

This increased model generality creates a corresponding systems challenge. Foundation models are task-independent where structured and unstructured data can be fed to it directly, without the feature engineering, schema design or label curation that once forced humans to inspect it. The generality gained on the model side is therefore accompanied by greater heterogeneity on the data side. Enterprise AI rarely operates on clean, homogeneous, model-ready inputs: organisational knowledge is authored and stored in the formats people use (PDFs, presentations, spreadsheets, scanned documents, forms, tables, diagrams, figures, images and mixed-layout files), and these artefacts may carry textual, visual, geometric and structural information simultaneously. These artefacts may contain textual, visual, geometric, and structural information simultaneously. A language model or retriever cannot reliably reason over information misrepresented at this interface. Thus, the generality gained on the model side is accompanied by greater heterogeneity on the data side. The engineering problem shifts from repeatedly constructing task-specific models to reliably transforming diverse enterprise content into representations that downstream models can process.

\begin{figure*}[t]
\centering
\includegraphics[width=\textwidth]{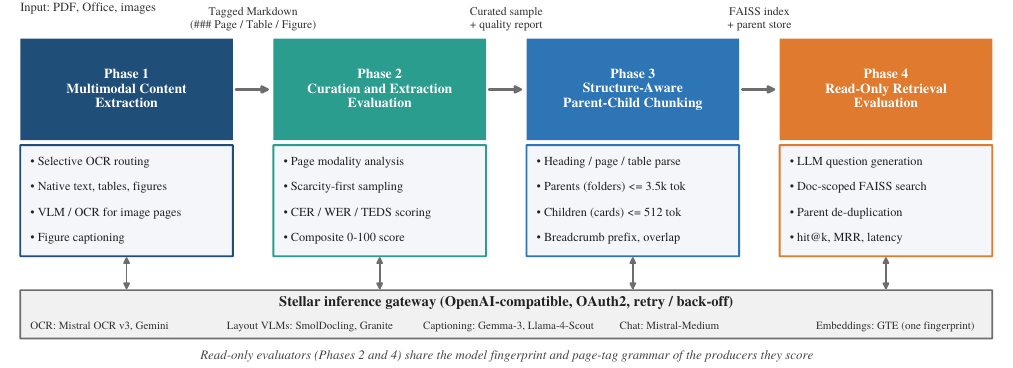}
\caption{Four phases turn raw files into a measured embeddings with a parent store. The embedding model used to index chunks and the one used to embed evaluation queries are byte-identical. The two evaluators (Phases 2 and 4) are strictly read-only.}
\label{fig:arch}
\end{figure*}

\textbf{This paper.} We describe the document-ingestion system that feeds \gls{GenAI} workloads. The system is organised as four phases that mirror the lifecycle of a document inside a RAG system (Figure~\ref{fig:arch}): multimodal content extraction, data curation and extraction evaluation, structure-aware parent-child chunking, and retrieval evaluation. Its guiding commitment is that \emph{each phase must produce evidence about itself}: extraction is scored against curated ground truth, chunking is scored by a retrieval examiner, and both evaluators are read-only and share the exact model fingerprints of the producers they score.

\textbf{Contributions.} In the spirit of the MLSys industrial track we report design methodology, implementation and detailed benchmarks of a production system rather than a new algorithm:
\begin{itemize}\setlength{\itemsep}{1pt}
\item A formalisation of \emph{selective OCR routing} and of structure-first native extraction, with a page-level router that sends only image-dominant pages to vision models (Section~\ref{sec:extract}).
\item A \emph{scarcity-first multi-label sampling} algorithm that builds modality-balanced evaluation sets from text-dominated corpora, and a reference-based scorer combining CER, WER and tree-edit-distance table similarity into a coverage-penalised composite (Section~\ref{sec:curate}).
\item A deterministic \emph{parent-child chunker} with four cutting rules, a safety overlap and contextual breadcrumbs, whose query-time context expansion is $O(1)$ (Section~\ref{sec:chunk}).
\item A \emph{read-only retrieval examiner} that generates grounded questions from every page and reports hit@$k$, MRR and latency per document, per question type and overall (Section~\ref{sec:eval}).
\item Benchmarks of four production OCR back-ends and of the full pipeline, with latency decomposition (Section~\ref{sec:results}), and lessons for building measurable ingestion for \gls{GenAI} systems.
\end{itemize}

Throughout, we label claims by evidence status: \E{} established prior research, \D{} documented measurements produced by the system's own evaluators, \Pp{} our proposed framework or design choice, and \Hh{} hypothesis or future direction.

\section{Background and Related Work}
\label{sec:related}

\paragraph{Document parsing and OCR.} \E{} Document AI has moved from engine-based OCR \cite{smith2007tesseract} and layout detectors \cite{zhong2019publaynet,shen2021layoutparser} to multimodal pre-training \cite{xu2020layoutlm,huang2022layoutlmv3} and OCR-free transformers \cite{kim2022donut,blecher2023nougat}. Open toolchains (Docling \cite{auer2024docling} and its compact SmolDocling and Granite-Docling variants \cite{nassar2025smoldocling}, MinerU \cite{wang2024mineru}, Marker \cite{marker}) convert PDFs to Markdown with layout models, while commercial vision models (Mistral OCR \cite{mistral2025ocr}, Gemini \cite{geminiteam2023}) offer page-to-Markdown as a service; OmniDocBench \cite{ouyang2025omnidocbench} is a public benchmark with page-level modality annotations. Table structure recognition has its own datasets and metrics \cite{gobel2013icdar,gao2019ctdar,li2020tablebank,smock2022pubtables}, of which the Encoder-Dual-Decoder model and the TEDS metric of \citet{zhong2020edd} matter most here; earlier work by one of the authors applied CNN table detection and EDD cell extraction to query-based table retrieval \cite{tufail2024tables}. The system does not propose a new parser; \Pp{} its contribution is the routing, scoring and auditing scaffold around existing ones (Table~\ref{tab:landscape}).

\paragraph{Chunking for retrieval.} \E{} Retrieval granularity is a first-order determinant of RAG quality \cite{chen2024densex}. Fixed-size and recursive splitters are cheap but structure-blind; semantic splitters place boundaries at embedding discontinuities; hierarchical schemes such as RAPTOR \cite{sarthi2024raptor} and small-to-big (parent-child) retrieval decouple what is matched from what is read; late chunking \cite{gunther2024latechunking} and contextual retrieval inject surrounding context into chunk embeddings. The approach adopts deterministic parent-child chunking driven by the document's own headings, page and table tags, and injects context through a breadcrumb prefix computed once at index time.

\paragraph{Evaluating RAG.} \E{} Component-wise frameworks (RAGAS \cite{es2024ragas}, ARES \cite{saadfalcon2024ares}) and benchmarks \cite{thakur2021beir,chen2024benchmarkingrag} measure retrieval and generation quality, often with LLM judges whose reliability must itself be validated \cite{zheng2023judge,shankar2024validators}. The system separates the two measurements that matter for ingestion, extraction fidelity (reference-based CER, WER, TEDS) and chunk retrievability (synthetic-question hit@$k$ and MRR \cite{craswell2009mrr}), and builds both as read-only harnesses.


\begin{table}[t]
\caption{Landscape of open document-ingestion tooling versus the proposed approach. $\bullet$ built in; $\circ$ partial or via plug-ins; - absent. Assessed from published descriptions \cite{auer2024docling,wang2024mineru,marker,blecher2023nougat,unstructured}.}
\label{tab:landscape}
\vskip 0.1in
\centering
\scriptsize
\setlength{\tabcolsep}{2.6pt}
\begin{tabular}{@{}lcccccc@{}}
\toprule
 & Docling & MinerU & Marker & Nougat & Unstr. & \textbf{This Paper} \\
\midrule
Native-text fast path & $\bullet$ & $\circ$ & $\bullet$ & - & $\bullet$ & $\bullet$ \\
Page-level OCR routing & $\circ$ & $\circ$ & $\circ$ & - & $\circ$ & $\bullet$ \\
Pluggable OCR back-ends & $\circ$ & $\circ$ & $\circ$ & - & $\circ$ & $\bullet$ \\
Structure-aware chunker & $\bullet$ & - & - & - & $\bullet$ & $\bullet$ \\
Balanced curation & - & - & - & - & - & $\bullet$ \\
Extraction scorer & - & - & - & - & - & $\bullet$ \\
Retrieval examiner & - & - & - & - & - & $\bullet$ \\
\bottomrule
\end{tabular}
\end{table}

\section{Problem Formulation and Design Principles}
\label{sec:problem}

\paragraph{Notation.} A corpus is a set of documents $\mathcal{D}$; document $d$ has pages $p \in P(d)$. Each page carries a text layer $T(p)$ (possibly empty) and a set of embedded images with total area $A_{\mathrm{img}}(p)$ on a page of area $A(p)$. An extractor $f_\theta$ maps a document to tagged Markdown $M = f_\theta(d)$ in which page boundaries, headings, tables and figures are explicit tags. A chunker $g_\phi$ maps $M$ to a set of child chunks $\mathcal{C}$ and parents $\mathcal{P}$ with a surjection $\mathrm{par}: \mathcal{C} \to \mathcal{P}$. An embedding model $e(\cdot)$ with fingerprint $\mathcal{F}$ maps text to $\mathbb{R}^n$. A query $q$ with source page $\pi(q)$ is answered by ranking $\mathcal{C}$ under cosine similarity.

\paragraph{Four coupled problems.} The system addresses, in order, (i) \emph{extraction}: choose $f_\theta$ and its routing so that $M$ is faithful to $d$ at minimal cost; (ii) \emph{curation}: select a sample $S$ of pages on which faithfulness can be measured with confidence; (iii) \emph{chunking}: choose $g_\phi$ so that chunks are both retrievable and sufficient; (iv) \emph{evaluation}: measure whether (iii) holds without mutating the index. The phases are coupled by shared artefacts (the tag grammar, the fingerprint $\mathcal{F}$), not by shared code paths.

\paragraph{Design principles.} Three principles recur across all four phases.
\begin{enumerate}\setlength{\itemsep}{1pt}
\item \textbf{Structure before semantics.} Use the document's own authored structure (font hierarchy, ruling lines, headings, page breaks) before invoking learned models; this is deterministic, cheap and debuggable.
\item \textbf{Never mutate what you measure.} Evaluators are read-only and reuse the producer's exact models, tokenizer and tag grammar, so an evaluation can never measure a different system from the one in production \cite{breck2017mltestscore,amershi2019se4ml}.
\item \textbf{Measure before you scale, and budget your labels.} Exhaustive ground truth is infeasible; a small, modality-balanced, traceably sampled set stands in for population-level confidence \cite{neyman1934,cochran1977sampling}.
\end{enumerate}

\section{Phase 1: Multimodal Content Extraction}
\label{sec:extract}

\subsection{Selective OCR routing}
\label{sec:routing}

\Pp{} The central efficiency principle is that the document connector extracts all directly accessible content without invoking OCR and routes only image-bound content to vision models. For each page the router computes the image-area ratio
\begin{equation}
r(p) = \frac{A_{\mathrm{img}}(p)}{A(p)}, \qquad
\mathrm{route}(p) = \begin{cases} \text{OCR} & r(p) > \tau_{\mathrm{cov}} \\ \text{native} & \text{otherwise,} \end{cases}
\label{eq:route}
\end{equation}
with $\tau_{\mathrm{cov}} = 0.8$ by default. Pages routed to OCR are rasterised at a configurable DPI; all other pages are parsed through PyMuPDF's structured text, table and image extraction \cite{pymupdf}. The curation subsystem (Section~\ref{sec:curate}) uses a stricter two-signal definition of a \emph{scanned} page, $s(p) = \ind{|T(p)| < \kappa \wedge r(p) > \tau_{\mathrm{cov}}}$ with $\kappa = 40$ characters, so that a page with a large photograph and a full text layer is not mislabelled as scanned. On the Orion corpus 853 of 10,507 pages (8.1\%) take the OCR route \D{}; the native path therefore carries 92\% of pages at near-zero model cost.


\subsection{Native-text structural reconstruction}
\label{sec:native}

For text pages, \code{PDFExtractor} reconstructs structure without any learned model.

\textbf{Font calibration.} Let $f^\ast$ be the modal font size over the document, treated as body text. Distinct sizes $f \ge \rho f^\ast$ (default $\rho = 1.12$) are sorted in descending order and assigned heading tiers $1, 2, \dots$ up to a configured depth $L$, yielding a heading-level function $h(f)$ used to emit Markdown headings.

\textbf{Header and footer suppression.} Text blocks lying within the page margins are keyed by their digit-normalised content (so that ``Page 3'' and ``Page 7'' collapse to one key); keys recurring on more than $\theta_{\mathrm{rep}} = 50\%$ of pages are treated as running furniture and dropped.

\textbf{Two-tier table detection.} Ruled tables are found by PyMuPDF's \code{find\_tables} with the \code{lines\_strict} strategy. When \code{tables\_strategy = auto}, a word-alignment detector additionally recovers borderless tables: words are clustered into visual rows, rows are grouped into columns by an $x$-alignment tolerance, and a guard (\code{looks\_like\_multi\_column\_text}) rejects side-by-side prose such as slide layouts. Detected tables are rendered as pipe tables tagged \code{\#\# Table-\{page\}.\{n\}}.

\textbf{Line-granularity reconciliation.} Rather than dropping any text block that overlaps a table, each line is tested individually: a line is suppressed only if more than 50\% of its area lies inside a table rectangle, preserving sentences that merely share a block with a table fragment.

\textbf{Figures, reading order and inline structure.} Embedded rasters above 32$\times$32\,px are de-duplicated by MD5 hash and replaced by a \code{\#\# Figure-\{page\}.\{n\}} placeholder. Blocks are ordered by $(y, x)$, optionally banded by column first. Bullets are normalised, monospace runs become fenced code, wrapped paragraphs are de-hyphenated, and hyperlinks are reconstructed from link annotations overlapping a span by at least 50\%.

\subsection{Image-based extraction}
\label{sec:ocr}

Image-dominant pages pass through an optional enhancement pipeline (border removal; denoising chosen among non-local-means \cite{buades2005nlm}, bilateral and Gaussian filters from a Laplacian-variance noise estimate; CLAHE \cite{zuiderveld1994clahe}; unsharp masking; cubic upscaling) and are sent to one of the available OCR language models. A retry and back-off loop handles transient 429/5xx responses and refreshes OAuth2 tokens on 401.


\subsection{Captioning, format coverage and output}

Any remaining \code{Figure} placeholder triggers a vision call whose caption is typed and substituted in place; tables inside OCR output are normalised and tagged so the chunker can treat them atomically. Word, PowerPoint, Excel and CSV files go to dedicated extractors and raw images directly into the OCR path, giving one format-agnostic entry point. The final Markdown, still carrying \code{\#\#\# Page: N} tags, is the single artefact consumed by both the extraction scorer and the chunker.

\section{Phase 2: Curation and Extraction Evaluation}
\label{sec:curate}

Phase 2 answers a prerequisite question: on what data, and with what confidence, do we know the extractor is accurate? It comprises a sampling engine, a reference-based scorer and a documented reference-free monitoring rationale.

\subsection{Page modality analysis}

Every discovered PDF page is analysed into a \code{PageRecord} with a multi-label modality vector $y_p \in \{0,1\}^M$ over the classes in Table~\ref{tab:modality}. Text pages use geometric heuristics; scanned pages are rasterised and sent to Mistral OCR with \code{annotate=True}, whose structured region typing is mapped onto the same label set. \Pp{} Reusing the OCR back-end's own annotation avoids a second heuristic pass and keeps the curation labels consistent with what the extractor actually sees.

\begin{table}[t]
\caption{Modality taxonomy and detection signals used by the curation engine.}
\label{tab:modality}
\vskip 0.1in
\centering
\small
\begin{tabular}{@{}lp{4.9cm}@{}}
\toprule
Class & Signal \\
\midrule
Scanned & $|T(p)| < \kappa$ and $r(p) > \tau_{\mathrm{cov}}$ \\
Text & non-scanned page with extractable text \\
Table & \code{page.find\_tables()} non-empty, or OCR \code{has\_table} \\
Diagram & vector drawing count above threshold, evaluated only when the page is not already a table (avoids counting gridlines) \\
Graph / infographic & image-area ratio inside a band $[a_{\min}, a_{\max}]$ that excludes icons and full-bleed photographs \\
Photo, signature & from OCR region annotation only \\
\bottomrule
\end{tabular}
\end{table}

\subsection{Scarcity-first representative sampling}
\label{sec:sampling}

\E{} Naive random sampling of a text-dominated corpus starves rare modalities: if 90\% of pages are plain text, a random subset over-represents text and leaves scanned pages, diagrams and infographics with too few examples to estimate error. Manual curation does not scale and is inconsistent between curators. The classical remedy is stratified sampling \cite{neyman1934,cochran1977sampling}, but pages carry \emph{multiple} labels, which makes exact stratification a multi-label problem \cite{sechidis2011multilabel}.

\begin{figure}[t]
\centering
\includegraphics[width=\columnwidth]{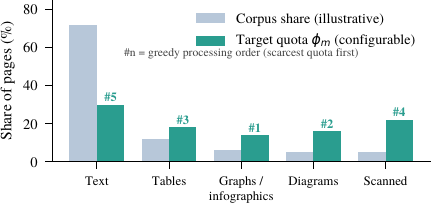}
\caption{Scarcity-first sampling. Target quotas $\phi_m$ are set independently of the corpus distribution; the greedy pass visits the scarcest quota first so that the dominant text class cannot exhaust the page budget. Corpus shares shown are illustrative.}
\label{fig:sampling}
\end{figure}

\Pp{} Given target fractions $\phi_m$ and a total budget $N$, quotas are $t_m = \lceil \phi_m N \rceil$, and the goal is a set $S$ with $|S| = N$ satisfying $\sum_{p \in S} y_{p,m} \ge t_m$ for all $m$. This is a covering problem with multiplicities and is NP-hard in general; the proposed system uses a greedy heuristic in the spirit of \citet{chvatal1979greedy} with two refinements (Algorithm~\ref{alg:sampling}, Figure~\ref{fig:sampling}): modalities are processed in ascending order of quota (scarcest first) so that abundant classes cannot exhaust the budget, and within a modality pages are ranked by a \emph{helpfulness} score, the number of other still-unmet quotas the page also satisfies.

\begin{algorithm}[t]
\caption{Scarcity-first multi-label sampling}
\label{alg:sampling}
\begin{algorithmic}[1]
\STATE \textbf{Input:} pages $\mathcal{U}$ with labels $y_p$; quotas $t_m$; budget $N$
\STATE $S \leftarrow \emptyset$; deficit $d_m \leftarrow t_m$ for all $m$
\FOR{$m$ in $\arg\mathrm{sort}_{\uparrow}(t_m)$}
  \STATE $\mathcal{A} \leftarrow \{p \in \mathcal{U}\setminus S : y_{p,m} = 1\}$
  \STATE sort $\mathcal{A}$ by $\mathrm{help}(p) = \sum_{m' \ne m} y_{p,m'} \ind{d_{m'} > 0}$ descending
  \WHILE{$d_m > 0$ and $\mathcal{A} \ne \emptyset$}
    \STATE pop $p$ from $\mathcal{A}$; $S \leftarrow S \cup \{p\}$; $d_{m'} \leftarrow d_{m'} - y_{p,m'}$ for all $m'$
  \ENDWHILE
\ENDFOR
\STATE top up with unselected pages or trim lowest-helpfulness pages until $|S| = N$
\STATE \textbf{Output:} $S$ and a Target\_vs\_Achieved audit sheet
\end{algorithmic}
\end{algorithm}

\textbf{Export and audit trail.} Selected pages are exported either as cropped single-page PDFs (lean evaluation sets) or as whole source PDFs (when document context is needed), preserving folder hierarchy, together with a six-sheet workbook (target versus achieved, input and output statistics, all and selected pages, configuration snapshot) that makes every sampling decision reproducible. Ground-truth digitisation and human validation are performed on this sample.

\subsection{Reference-based extraction scoring}
\label{sec:scoring}

Given ground truth $c$ and hypothesis $\hat c$ for a file, the scorer first normalises both (Unicode NFKC, case folding, punctuation and ligature unification, Markdown and HTML stripping, whitespace collapsing) so that formatting differences between a plain-text reference and Markdown output are not counted as content errors. It then computes
\begin{equation}
\mathrm{CER} = \frac{\mathrm{Lev}_{\mathrm{char}}(\hat c, c)}{|c|}, \qquad
\mathrm{WER} = \frac{\mathrm{Lev}_{\mathrm{word}}(\hat c, c)}{|c|_{\mathrm{w}}},
\end{equation}
using Levenshtein distance \cite{levenshtein1966}, together with character- and word-level precision, recall and $F_1$, an order-insensitive bag-of-words $F_1$, smoothed corpus BLEU-4 \cite{papineni2002bleu}, and a length ratio $|\hat c| / |c|$ that flags over-production (duplication or hallucination \cite{ji2023hallucination}) or content loss.

\textbf{Table structure.} Pipe tables and HTML tables (with colspan and rowspan expansion) are parsed into rectangular grids and compared with TEDS \cite{zhong2020edd}: tables are modelled as $\mathrm{table} \to \mathrm{row}^\ast \to \mathrm{cell}^\ast$ trees and
\begin{equation}
\mathrm{TEDS}(T_a, T_b) = 1 - \frac{\mathrm{EditDist}(T_a, T_b)}{\max(|T_a|, |T_b|)},
\end{equation}
where the row-sequence edit distance uses, as substitution cost, a cell-sequence edit distance whose own substitution cost is the normalised Levenshtein distance between cell strings (zero in \code{structure\_only} mode) \cite{pawlik2016ted}.

\textbf{Composite score and ranking.} Both corpus-weighted (micro) and per-file (macro) aggregates are computed. The composite used to rank back-ends is
\begin{equation}
Q_m = 100 \cdot \frac{\sum_j w_j\, \mu_{j,m}}{\sum_j w_j} \cdot \mathrm{cov}_m,
\label{eq:composite}
\end{equation}
where $\mu_{j,m}$ is model $m$'s mean on accuracy-oriented metric $j$ (character accuracy, word accuracy, word $F_1$, bag-of-words $F_1$, BLEU, table TEDS, table detection $F_1$, table cell $F_1$), $w_j$ are configurable weights, and $\mathrm{cov}_m$ is the fraction of expected files the model actually produced. \Pp{} The coverage factor explicitly penalises a model that ``wins'' by skipping hard documents. Each ranked model is annotated with human-readable verdicts (``near-perfect text'', ``weak tables'', ``over-produces text'') derived from threshold bands on CER, TEDS and length ratio.

\subsection{Reference-free monitoring rationale}

\Hh{} CER and WER require references that production traffic rarely has. A reference-free composite (50\% lexical validity, 30\% character sanity, 20\% token shape) can rank pages without labels but cannot report a true error rate. Three complementary calibration strategies are documented as rationale rather than shipped code: synthetic ground truth (known text rendered to degraded images, for CI regression), engine-consensus pseudo-ground-truth (two independent engines agreeing, for continuous monitoring), and small calibration sets (10 to 20 hand-corrected pages that anchor the reference-free score to true CER, for alerting).

\section{Phase 3: Structure-Aware Parent-Child Chunking}
\label{sec:chunk}

\subsection{The granularity problem}

\E{} After extraction, chunk sizing is the most consequential design decision \cite{chen2024densex}, and both extremes fail characteristically. Too large, and the embedding averages over all content: under a mean-pooling approximation $e(c) \approx \frac{1}{n}\sum_i e(s_i)$ over $n$ sentences, one relevant sentence's contribution to $\cos(e(q), e(c))$ shrinks with $n$ and the correct chunk may never surface (\emph{score dilution}). Too small, and a complete answer spans several fragments none of which scores highly alone (\emph{score fragmentation}). 

\subsection{Small cards, big folders}

This system resolves the tension by decoupling the \emph{retrieval unit} from the \emph{context unit}. Children are small, precisely embedded index cards (target 320 tokens, never more than 512) stored in a FAISS index; parents are the folders they live in (up to 3,500 tokens) stored in a key-value map from \code{parent\_id} to text. The card-to-folder link is established once at indexing time, so query-time context expansion is one $O(1)$ lookup. Table~\ref{tab:chunkers} positions this against alternative strategies.

\begin{table}[t]
\caption{Chunking strategies on properties that matter for auditable ingestion: fixed and recursive splitters, semantic splitting, RAPTOR \cite{sarthi2024raptor}, late chunking \cite{gunther2024latechunking} and contextual retrieval \cite{anthropic2024mcp}.}
\label{tab:chunkers}
\vskip 0.1in
\centering
\small
\setlength{\tabcolsep}{3.5pt}
\begin{tabular}{@{}lllll@{}}
\toprule
Strategy & Det. & Structure & Index & Query \\
\midrule
Fixed / recursive & yes & none & low & low \\
Semantic & no & none & high & low \\
RAPTOR & no & summaries & v.\ high & medium \\
Late chunking & yes & implicit & high & low \\
Contextual retrieval & no & LLM prefix & v.\ high & low \\
Proposed parent-child & yes & explicit tags & low & $O(1)$ \\
\bottomrule
\end{tabular}
\end{table}


\subsection{Parsing and parent construction}

The Phase-1 Markdown is parsed line by line with three matchers: Markdown headings, a page tag built from the configured \code{page\_tag\_pattern}, and table and figure tags. Page tags are stripped but their number is attached to every subsequent block; headings push and pop a stack of (level, title) that becomes each block's \code{section\_path}; headings matching the table or figure pattern open an \emph{atomic block} rather than a structural level. Table bodies are tracked row by row so that prose following a table on the same page is split back into text, and a page tag inside an open atomic block decides whether a continuation is still the same table or has become prose. Parents are formed by greedily packing a section's blocks, never splitting one, into groups bounded by \code{parent\_min\_tokens} and \code{parent\_max\_tokens}; a table or figure that alone exceeds the limit is kept whole.

\subsection{Children: four cutting rules, overlap and breadcrumbs}

Children are built by \code{\_pack\_sentences()} under four rules (Algorithm~\ref{alg:children}): (R1) target about 320 tokens and never exceed 512; (R2) never cut mid-sentence, using a punctuation-aware splitter that keeps a trailing quote or bracket with its sentence, and emitting an over-long single sentence as its own chunk rather than truncating; (R3) no crumbs: a packed group below \code{child\_min\_tokens} is merged into the previous chunk's non-overlap sentences if the merge respects the ceiling; (R4) tables and figures are atomic, each flushing any pending prose and becoming exactly one child. When a chunk closes because the next sentence would exceed the ceiling, up to \code{overlap\_tokens} (default 48, roughly one or two sentences) of trailing sentences are seeded into the next chunk so that an answer straddling a boundary appears intact on at least one card.

\textbf{Contextual breadcrumbs.} When enabled, each child's \code{embed\_text} is prefixed with its structural address, \code{Document > Title > Section > Subsection (p.N)}, so that the vector encodes both meaning and position. This is the deterministic analogue of contextual retrieval at zero LLM cost; the unlabelled \code{raw\_text} is preserved for display. Each child carries \code{chunk\_id, doc\_id, parent\_id, kind, section\_path, page\_id, raw\_text, embed\_text, token\_count}; each parent carries its id, document, section path, page span, text, token count and child ids. Tokens are counted with the GPT-2 tokenizer \cite{radford2019gpt2}, the same tokenizer used by the evaluator.

\begin{algorithm}[t]
\caption{Child packing for one parent}
\label{alg:children}
\begin{algorithmic}[1]
\STATE \textbf{Input:} blocks $B$; bounds $(\ell_{\min}, \ell_{\mathrm{target}}, \ell_{\max})$; overlap $o$
\STATE buffer $\leftarrow []$; carry $\leftarrow []$
\FOR{block $b \in B$}
  \IF{$b$ is table or figure}
    \STATE flush(buffer) as a child; emit $b$ as one atomic child \hfill (R4)
  \ELSE
    \FOR{sentence $s$ in split($b$)}
      \IF{tok(buffer) + tok($s$) $> \ell_{\max}$}
        \STATE emit child(carry $\Vert$ buffer); carry $\leftarrow$ last sentences of buffer with $\le o$ tokens
        \STATE buffer $\leftarrow [s]$ \hfill (R1, R2)
      \ELSE
        \STATE append $s$ to buffer
      \ENDIF
    \ENDFOR
  \ENDIF
\ENDFOR
\STATE flush(buffer); merge any child below $\ell_{\min}$ into its predecessor if the result $\le \ell_{\max}$ \hfill (R3)
\end{algorithmic}
\end{algorithm}

\subsection{Embedding, indexing and auditable outputs}

Children are embedded in batches of 32 against GTE and L2-normalised so that FAISS \code{IndexFlatIP} inner product equals cosine similarity \cite{douze2024faiss}. Indexing is incremental and resumable: an existing index and metadata pair is loaded, extended under a lock, and rewritten as one consolidated index plus JSON metadata, so the corpus is never re-embedded wholesale.

\section{Phase 4: Read-Only Retrieval Evaluation}
\label{sec:eval}

\begin{figure}[t]
\centering
\includegraphics[width=\columnwidth]{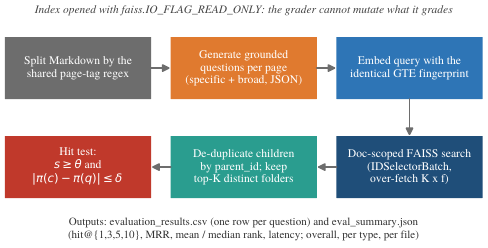}
\caption{The retrieval examiner. Page splitting reuses the chunker's \code{build\_page\_regex()}; query embedding reuses the chunker's model fingerprint; search is scoped to the source document; hits are judged on score and page proximity.}
\label{fig:evalflow}
\end{figure}

\Pp{} \code{RetrievalEvaluator} is designed solely for measurement, not modification: it opens the index in read-only mode (Figure~\ref{fig:evalflow}).

\textbf{Question generation.} Pages are recovered from the chunker's Markdown with the identical page-tag regex. For each page with at least \code{min\_page\_tokens} tokens, a chat model (Mistral-Medium) produces up to $n_s$ \emph{specific} questions (fact-level, answerable from one passage, table or figure) and $n_b$ \emph{broad} questions (synthesis across the page), grounded strictly in the page text under a prompt that forbids outside knowledge and enforces raw JSON. Grounded synthetic questions are a reference-free proxy in the tradition of ARES and RAGAS \cite{saadfalcon2024ares,es2024ragas}, with the known caveat that their quality depends on the generator \cite{zheng2023judge}.

\textbf{Query execution and hit criterion.} Each question is embedded with the same GTE fingerprint $\mathcal{F}$ (a strict requirement for vector comparability) and searched within its source document via an \code{IDSelectorBatch} over that document's rows, with an exhaustive-filter fallback for FAISS builds lacking selector support so that in-document results are never silently dropped. Retrieved children are de-duplicated by \code{parent\_id}, keeping the highest-scoring child per parent, after over-fetching by a factor $f$ so that a full list of $K$ distinct parents survives. With $s(q,c) = \cos(e(q), e(c))$, candidate $c$ is a hit for question $q$ if
\begin{equation}
\mathrm{hit}(q, c) = \ind{s(q,c) \ge \theta \,\wedge\, |\pi(c) - \pi(q)| \le \delta},
\end{equation}
with page tolerance $\delta = 1$ to accommodate answers that straddle a page boundary. Over the question set $\mathcal{Q}$ with $\mathrm{rank}_q$ the first hit rank,
\begin{equation}
\mathrm{hit@}k = \tfrac{1}{|\mathcal{Q}|}\textstyle\sum_{q} \ind{\mathrm{rank}_q \le k}, \;\;
\mathrm{MRR} = \tfrac{1}{|\mathcal{Q}|}\textstyle\sum_{q} \mathrm{rank}_q^{-1},
\end{equation}
following standard definitions \cite{craswell2009mrr}. Per-query latency (embedding plus search) is recorded for every question. Evaluation is parallelised across documents with a thread pool (I/O-bound, since each query is a remote embedding call); outputs are one CSV row per question and a JSON summary with hit@$\{1,3,5,10\}$, MRR, mean and median rank, and mean latency, broken down overall, per question type and per file.

\section{Results}
\label{sec:results}

All numbers in this section are \D{} artefacts produced by the system's own evaluators on internal corpora (``icm'', ``orion'', ``polaris'', ``uspb''); the Orion corpus comprises 180 documents, 9,654 text pages and 853 image pages. They are the system's self-measurements, which is the auditability property the architecture exists to deliver.

\subsection{Latency decomposition}

\begin{figure}[t]
\centering
\includegraphics[width=\columnwidth]{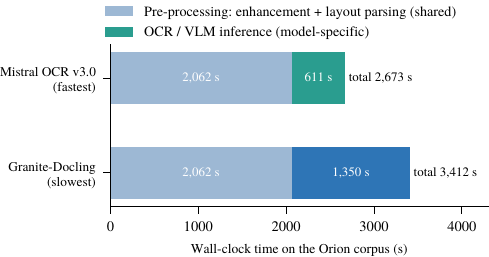}
\caption{Wall-clock decomposition for the Orion corpus (180 documents, 9,654 text and 853 image pages). The shared pre-processing stage dominates; the two other back-ends fall between the extremes shown.}
\label{fig:latency}
\end{figure}

On Orion, end-to-end extraction took between 2,672.67\,s (Mistral OCR) and 3,411.86\,s (Granite-Docling). Pre-processing (image enhancement and layout parsing) is a fixed cost of roughly 2,062\,s shared by all back-ends; model inference is the differentiator at roughly 0.6 to 1.35 thousand seconds (Figure~\ref{fig:latency}). Two consequences follow: with 92\% of pages on the native route the model-dependent share of wall-clock time is already a minority, so the next optimisation target is the shared pre-processing stage; and the fastest model is not the most accurate, so back-end choice is a cost-quality trade-off that the composite score and benchmarking sheet make explicit.

\subsection{Extraction quality}

\begin{figure*}[t]
\centering
\includegraphics[width=\textwidth]{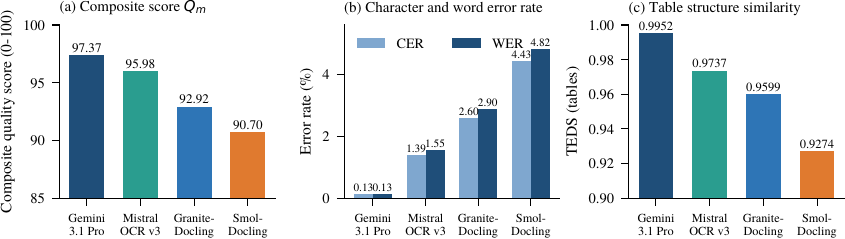}
\caption{Extraction benchmark across four back-ends: (a) composite score, (b) character and word error rates, (c) table structure similarity (TEDS). Text and table fidelity rank the models consistently.}
\label{fig:extraction}
\end{figure*}

Table~\ref{tab:extraction} and Figure~\ref{fig:extraction} rank the four production back-ends by Equation~\ref{eq:composite}. Gemini 3.1 Pro leads with 97.37 and a CER of 0.13\%; Mistral OCR v3 follows at 95.98 with an order of magnitude higher CER (1.39\%) but TEDS still above 0.97. The two compact Docling models trail by 4 to 7 points, with table structure (TEDS 0.93 to 0.96) accounting for more of the gap than text. Prior EDD-based cell extraction reported 85\% word accuracy on image tables \cite{tufail2024tables}; the production back-ends here operate at error rates one to two orders of magnitude lower, which illustrates how fast the extraction frontier has moved.

\begin{table}[t]
\caption{Extraction back-ends ranked by the coverage-penalised composite score (Equation~\ref{eq:composite}); lower CER and WER and higher TEDS are better.}
\label{tab:extraction}
\vskip 0.1in
\centering
\small
\begin{tabular}{@{}lrrrr@{}}
\toprule
Model & Score & CER & WER & TEDS \\
\midrule
Gemini 3.1 Pro & \textbf{97.37} & \textbf{0.0013} & \textbf{0.0013} & \textbf{0.9952} \\
Mistral OCR v3 & 95.98 & 0.0139 & 0.0155 & 0.9737 \\
Granite-Docling (258M) & 92.92 & 0.0260 & 0.0290 & 0.9599 \\
SmolDocling (256M) & 90.70 & 0.0443 & 0.0482 & 0.9274 \\
\bottomrule
\end{tabular}
\end{table}

\subsection{Retrieval quality of the chunking}

\begin{figure}[H]
\centering
\includegraphics[width=\columnwidth]{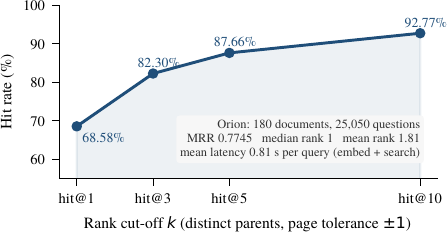}
\caption{Hit rate as a function of the rank cut-off on the Orion corpus. The steep rise from hit@1 to hit@3 and the median rank of 1 indicate that the correct parent is usually retrieved at or near the top.}
\label{fig:retrieval}
\end{figure}

On Orion the examiner generated and evaluated 25,050 questions, of which 1,810 missed at all ranks (Table~\ref{tab:retrieval}, Figure~\ref{fig:retrieval}). More than two thirds of questions return the correct parent as the very first result (hit@1 = 68.58\%); 92.77\% are answered within ten distinct parents; MRR is 0.7745 with a median rank of 1 and a mean rank of 1.81, indicating that misses are concentrated in a long tail rather than spread across the distribution. Mean query latency, embedding plus search over a flat index, is 0.81\,s and is dominated by the remote embedding call.

\begin{table}[t]
\caption{Retrieval evaluation on Orion (GTE embeddings, FAISS flat index, parent de-duplication, page tolerance $\pm 1$).}
\label{tab:retrieval}
\vskip 0.1in
\centering
\small
\begin{tabular}{@{}lr@{}}
\toprule
Metric & Value \\
\midrule
Documents / questions evaluated & 180 / 25,050 \\
Questions missed at all ranks & 1,810 (7.2\%) \\
hit@1 / 3 / 5 / 10 (\%) & 68.58 / 82.30 / 87.66 / 92.77 \\
MRR & 0.7745 \\
Median / mean rank & 1.0 / 1.81 \\
Mean per-query latency (s) & 0.81 \\
\bottomrule
\end{tabular}
\end{table}

A small offline rehearsal (two documents, 54 questions, mock fingerprints, threshold disabled) reported hit@1 of 70\%, hit@3 of 95\% and MRR 0.91; it exercised the harness before the production run and is not representative.

\section{Conclusion}
\label{sec:conclusion}

The proposed system shows that a document-ingestion pipeline for \gls{GenAI} systems can be made measurable end to end: every extraction, curation, chunking and retrieval decision is produced by an explicit, configurable and independently testable component, with read-only evaluators for both the extraction and the chunking layers. The reported figures (a 97.37-point extraction score for the best back-end; MRR 0.7745 and hit@1 of 68.58\% over 25,050 questions) are artefacts the system generates about itself. That self-measurement, more than any single algorithm, is what enterprise agentic systems will need most.

\bibliography{main}
\bibliographystyle{mlsys2025}

\end{document}